\documentclass[runningheads]{llncs}

\usepackage[T1]{fontenc}
\usepackage[utf8]{inputenc}
\usepackage{graphicx}
\usepackage{amsmath}
\usepackage{booktabs}
\usepackage{array}
\usepackage{url}
\usepackage[hidelinks]{hyperref}
\begin{document}

\title{Improved Vision-to-Chart Buoy Association with Learned World-to-Image Projection}
\titlerunning{QueryMLP for Vision-to-Chart Association}
\author{Borja Carrillo-Perez}
\authorrunning{B. Carrillo-Perez}

\institute{Arquimea Research Center\\
\email{bcarrillo@arquimea.com}}

\maketitle

\begin{abstract}
This report presents a lightweight modification to the DETR-based fusion transformer baseline for the MaCVi 2026 Vision-to-Chart data association challenge. The challenge baseline decoder receives per-buoy queries encoding world-space distance and bearing, forcing the transformer to implicitly learn the complex geometric projection from world coordinates to image pixels. Instead, in this work, an additional dedicated MLP (QueryMLP) is trained to explicitly predict the buoy's waterline contact point in the image from chart measurements and IMU orientation data. The predicted pixel coordinates are appended to the query vector of the baseline decoder, providing a direct spatial prior per buoy and reducing the geometric reasoning burden on the transformer decoder. On the challenge leaderboard the presented approach achieves an Overall score of $0.7386$ ($F_1 = 0.8055$, $\mathrm{mIoU} = 0.6718$) on the held-out test set, placing second among all submissions. Code is available at \url{https://github.com/bcarrpe/macvi26-visionmap-querymlp}.
\end{abstract}

\section{Introduction}

The MaCVi 2026 Vision-to-Chart challenge\footnote{\url{https://macvi.org/workshop/cvpr/challenges/vision_map}} requires detecting buoys in maritime camera images and matching each detection to its corresponding entry in a nautical chart. Each chart entry provides the buoy's GPS-derived distance and bearing relative to the vessel, while an Inertial Measurement Unit (IMU) file records pitch, roll, heading, and position at capture time. The evaluation metric is the mean of $F_1$-score and $\mathrm{mIoU}$ (Overall).

The challenge baseline~\cite{kreis2025realtime} modifies DETR~\cite{carion2020end} by replacing its fixed learned object queries with a variable number of chart-derived embeddings. Per-buoy features (normalized distance and bearing) are passed through a small multilayer perceptron (MLP) to produce $256$-dimensional decoder inputs. The transformer decoder then cross-attends these embeddings against ResNet-50 image features to predict visibility and bounding box location for each chart entry. Because the query encodes only world-space measurements, the decoder must implicitly learn the geometric projection from world coordinates to image pixels; this mapping depends on the vessel's instantaneous pitch, roll, and heading.

The presented approach addresses this directly, inspired by prior work of Carrillo-Perez et al. in maritime ship recognition and georeferencing~\cite{carrilloperez2022ship,carrilloperez2024thesis}. They demonstrate that a learned mapping can bridge world coordinates and pixel coordinates from paired training data alone, without explicit camera calibration. QueryMLP is trained to learn this world-to-image projection using distance, inverse of distance, bearing, and IMU orientation as inputs. Its output serves as a spatial prior that tells the decoder approximately where each buoy should appear in the image, thereby reducing what the decoder must discover from scratch to appearance-based refinement. A further insight from Carrillo-Perez et al.~\cite{carrilloperez2022ship,carrilloperez2024thesis} motivates the choice of prediction target: the waterline contact point of a maritime object is the most geometrically meaningful image-space location, since it corresponds directly to the object's position on the water plane. For buoys, this point is approximated by the bottom-center edge of the bounding box $[c_x,\; c_y + h/2]$, used as the MLP output.

\section{Method}

\subsection{QueryMLP: Learned World-to-Image Projection}

QueryMLP is trained independently of the main DETR pipeline to predict the normalized buoy waterline contact point from six features derived from the chart query and IMU files. Unlike the baseline, which forms decoder queries from normalized distance and bearing alone, QueryMLP also uses the available IMU orientation measurements to learn the world-to-image projection:

\begin{itemize}
    \item Normalized distance $(d / 1000)$: scale-invariant range prior.
    \item Inverse distance (clipped to $[0, 10]$): nonlinear depth cue that responds sharply to nearby buoys.
    \item Normalized bearing $(\beta / 180)$: horizontal image position prior.
    \item Pitch (normalized by $10^\circ$): shifts the expected vertical position with vessel pitch.
    \item Roll (normalized by $10^\circ$): accounts for lateral horizon tilt.
    \item Heading (normalized by $180^\circ$): global bearing context.
\end{itemize}

The normalization constants reflect the physical operating ranges of each quantity. Distance is divided by $1000$ m, the challenge's maximum range threshold, mapping the full operating range to $[0, 1]$. Inverse distance is clipped at $10$: without clipping, a buoy at $1$ m would yield $\mathrm{inv\_dist} = 1000$, dominating the network input; clipping at $10$ corresponds to distances of $100$ m or closer. This preserves the useful nonlinear variation at realistic ranges while preventing extreme values from destabilizing training. Bearing ranges from $-180^\circ$ to $+180^\circ$, so dividing by $180$ is standard unit normalization for this angular quantity. Pitch and roll are divided by $10^\circ$, which reflects the small vessel motion range observed in the dataset, mapping the actual data range to approximately $[-1, 1]$. Heading similarly ranges over $\pm 180^\circ$ and is normalized by $180$ for the same reason.

\begin{figure}[t]
\centering
\includegraphics[width=0.8\linewidth]{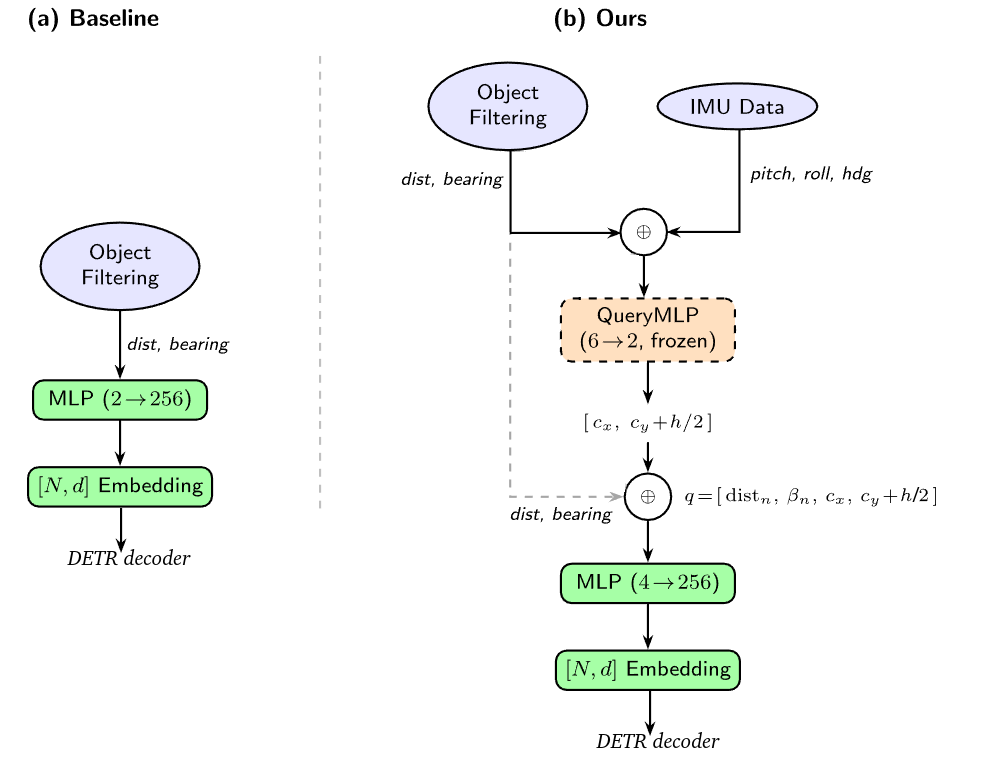}
\caption{Query construction pipeline. \textbf{(a)}~Baseline: chart distance and bearing are fed directly into the embedding MLP. \textbf{(b)}~Ours: a frozen QueryMLP takes six features (distance, bearing, and three IMU orientation angles) and predicts the buoy waterline contact point $[c_x, c_y\!+\!h/2]$ in image coordinates. These pixel coordinates are concatenated with the normalized distance and bearing to form a $4$-dimensional query vector, widening the embedding MLP input from $2$ to $4$. All other architecture components are unchanged from~\cite{kreis2025realtime}.}
\label{fig:arch}
\end{figure}

QueryMLP is a four-layer network with architecture $6 \rightarrow 128 \rightarrow 128 \rightarrow 128 \rightarrow 2$. Each hidden layer is followed by BatchNorm1d, ReLU, and Dropout$(0.2)$, with a Sigmoid output layer. The ground-truth target is the normalized horizontal center and bottom edge $(c_x, c_y + h/2)$, corresponding to the buoy's waterline contact point and extracted directly from the existing challenge annotations. It is trained on paired (query, IMU, label) examples from the training set using AdamW $(\mathrm{lr} = 10^{-3}$, weight decay $= 10^{-4})$ with batch size $256$, cosine annealing, and SmoothL1 loss, for up to $1000$ epochs with early stopping (patience $= 60$), converging at epoch $585$. QueryMLP is evaluated using the Euclidean pixel distance between its predicted point and the ground-truth target $(c_x,\; c_y + h/2)$. The model achieves on the validation set a median error of $18.4$ px (mean $= 27.1$ px, $90$th percentile $= 59.4$ px). Figure~\ref{fig:arch} illustrates the QueryMLP architecture in comparison with the baseline query construction.

\subsection{Integration into the DETR Pipeline}

Once trained, QueryMLP is held fixed. During DETR training and inference the dataset loader runs it on every chart query to produce predicted pixel coordinates $[c_x,\; c_y + h/2]$. They are then concatenated with the normalized distance and bearing to form a four-dimensional query vector
\[
q = [\mathrm{dist}_{\text{norm}}, \mathrm{bearing}_{\text{norm}}, c_x,\; c_y + h/2].
\]
This augmented query is then passed through the baseline decoder MLP, whose input dimension is widened from $2$ to $4$, to produce the $256$-dimensional decoder embedding used by DETR. QueryMLP is kept fixed during DETR training and inference. All other aspects of the architecture remain identical to the baseline: the ResNet-50 backbone, the six-layer transformer encoder and decoder, and the BCE $+$ L1 $+$ GIoU loss.

\section{Results}

The ATON dataset~\cite{kreis2025realtime}\footnote{\url{https://drive.google.com/drive/folders/1OXSok1Aux0rfygNQHFIe8goB695DeN3f}} contains $5{,}189$ samples ($4{,}285$ train / $904$ val) at $960 \times 540$ px. The method development and model selection were performed on the public train/validation split of the dataset. The final leaderboard results were obtained from the evaluation by the challenge organizers on a private held-out test set. Not all images contain visible buoys; empty label files indicate buoy-free frames. Performance is evaluated by $\mathrm{Overall} = (F_1 + \mathrm{mIoU}) / 2$.

The DETR model is initialized using COCO-pretrained DETR-R50 weights~\cite{carion2020end} and fine-tuned with AdamW (transformer $\mathrm{lr} = 10^{-4}$, backbone $\mathrm{lr} = 10^{-5}$) with auxiliary decoder losses, for approximately $185$ epochs with a StepLR drop at epoch $135$. The best validation checkpoint at epoch $182$ is the submitted solution.

At inference, a logit bias of $-0.5$ is applied to the raw predicted objectness scores before thresholding at $0.90$, following the challenge evaluation setup. The bias is selected by a grid sweep over the validation set (step $0.25$ in $[-3, 3]$) that evaluates Overall at each candidate value and picks the maximum. This calibration requires no retraining.

To provide a controlled reference point, the baseline architecture of~\cite{kreis2025realtime} is re-trained under identical conditions — same COCO initialisation, hyperparameters, augmentations, and number of epochs — but using only the original two-dimensional query (normalised distance and bearing), without IMU input or pixel coordinate prediction. The optimal logit bias for the baseline is found by the same grid sweep, yielding a bias of $-0.25$.

\begin{table}[!]
\caption{Validation results for the baseline and submitted solution, and test leaderboard result.}
\label{tab:results}
\centering
\begin{tabular}{lccccc}
\toprule
Model (split) & P & R & $F_1$ & mIoU & Overall \\
\midrule
Re-trained baseline (val) & 0.7970 & 0.7912 & 0.7941 & 0.6445 & 0.7193 \\
Ours (val) & 0.8627 & 0.7761 & 0.8171 & 0.6753 & 0.7462 \\
\midrule
Ours (test, 2nd place) & 0.8563 & 0.7604 & 0.8055 & 0.6718 & 0.7386 \\
\bottomrule
\end{tabular}
\end{table}

\begin{figure}[!h]
\centering
\includegraphics[width=\linewidth]{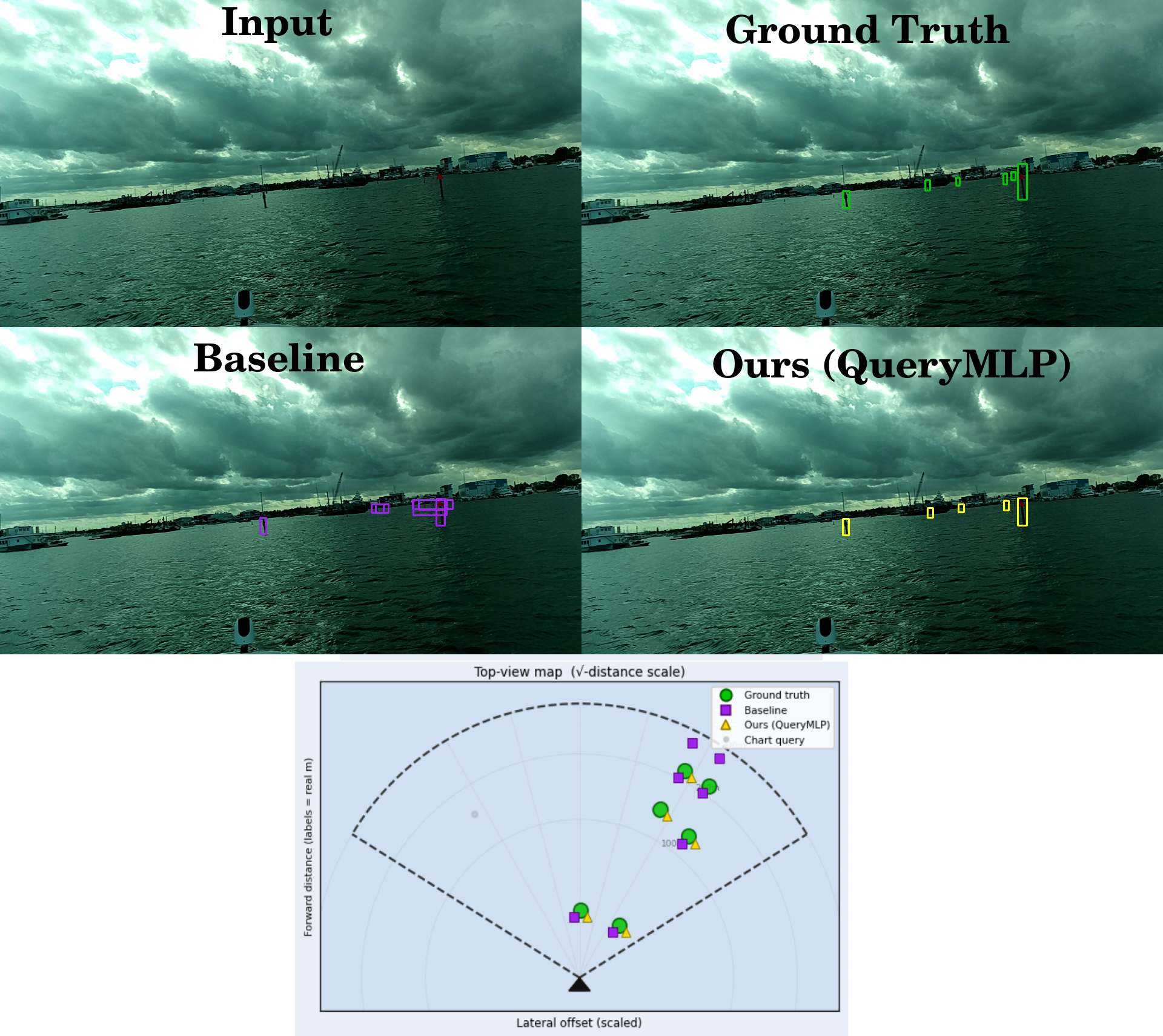}
\caption{Qualitative comparison on validation sample 00079. \textbf{Row 1}: input image and ground truth. \textbf{Row 2}: baseline predictions and ours. \textbf{Row 3}: top-view map of buoy positions ($\surd$-distance scale; green circle = ground truth, purple square = baseline, yellow triangle = ours). 
The baseline fires two false positive detections on non-buoy objects (left column, row 2); our method suppresses both while correctly detecting a buoy the baseline misses in this example, consistent with the improved precision and Overall score reported in Table~\ref{tab:results}.}
\label{fig:qual}
\end{figure}

Table~\ref{tab:results} shows the final scores on both the validation and held-out test sets. The validation and test Overall scores are consistent ($0.746$ vs. $0.739$), suggesting the approach generalizes to the test's unseen sequences. This submission ranked second on the private test leaderboard. Figure~\ref{fig:qual} shows a representative validation example illustrating the precision improvement.

\section{Conclusion}

This report presents a modification to the DETR-based vision-to-chart fusion transformer: a custom trained MLP that predicts the pixel coordinates of the buoy's waterline contact point from world-space chart measurements and IMU orientation data. Appending these coordinates to the existing query vector of the decoder provides the transformer with an explicit spatial prior per buoy, reducing the geometric reasoning it must learn from scratch. The approach achieves Overall $= 0.7386$ on the challenge test set (2nd place). Key design choices include the use of the buoy's waterline contact point as prediction target and inverse distance as a nonlinear depth cue. End-to-end training of QueryMLP jointly with the transformer decoder is a natural direction for future work, as the current frozen design prevents the decoder from providing feedback to refine the pixel predictions. The waterline contact point target, while geometrically stable in calm conditions, may be less reliable in rough seas where wave action or spray obscures the base of the buoy; these are precisely the conditions where robust vision-to-chart association is most safety-critical. Code is available at \url{https://github.com/bcarrpe/macvi26-visionmap-querymlp}.

\bibliographystyle{refs-style}
\bibliography{refs}

\end{document}